\documentclass[conference,flushend]{iaria}

\usepackage{amsmath,amssymb}
\usepackage{booktabs}
\usepackage{microtype}
\usepackage{siunitx}
\usepackage{xurl}
\usepackage{balance}
\usepackage{tabularx}
\usepackage{array}
\usepackage{placeins}

\usepackage{tikz}
\usetikzlibrary{positioning, arrows.meta, shapes.geometric}

\hypersetup{
  pdftitle={A Browser-Native Digital Test Range for Benchmarking 4D Ocean-Glider Planning Algorithms},
  pdfauthor={Edward Holmberg, Elias Ioup, and Mahdi Abdelguerfi}
}

\title{A Browser-Native Digital Test Range for Benchmarking 4D Ocean-Glider Planning Algorithms}
\author{
    \begin{tabular}[t]{c}
        \textbf{Edward Holmberg} \\
        Gulf States Center for \\
        Environmental Informatics \\
        LSU New Orleans, LA, USA \\
        eholmber@lsuneworleans.edu
    \end{tabular}
    \hfill
    \begin{tabular}[t]{c}
        \textbf{Elias Ioup} \\
        Ocean Sciences Division \\
        U.S. Naval Research Lab \\
        Stennis Space Center, MS, USA \\
        elias.z.ioup.civ@us.navy.mil
    \end{tabular}
    \hfill
    \begin{tabular}[t]{c}
        \textbf{Mahdi Abdelguerfi} \\
        Gulf States Center for \\
        Environmental Informatics \\
        LSU New Orleans, LA, USA \\
        gulfsceidirector@lsuneworleans.edu
    \end{tabular}
}

\begin{document}
\maketitle

\begin{abstract}
Repeated in-situ evaluation of ocean-glider planners requires scarce vehicles, operators, deployment and recovery resources, and ocean conditions that cannot be reset for competing algorithms. We present a guided, installation-free browser-native digital test range that transforms a selected region into a reproducible four-dimensional experiment. The system leads users from regional domain selection through mission-scoped bathymetry, time/depth forcing, science objectives, optional task decomposition, route specification, current-advected execution, observation generation, and scoring. Its primary contribution is a common plan-to-observation contract unifying vehicle, sensing, and evaluator assumptions across manual routes, transparent built-in algorithms, and imported classical or learned-planner outputs, while exported artifacts form dataset-ready records. A controlled Observing System Simulation Experiment (OSSE) evaluates five classical planners in two episodes, three deterministic seeds, and a calibrated 60-hour horizon. All 54 missions completed and recovered without hard violations, while planner rankings and dive-policy effects revealed operational-scientific tradeoffs. An authentic public deployment supplied a field-referenced audit to scope current kinematic boundaries. Separately, source-locked GliderFlight 1.2.0 achieved native-to-browser parity through Pyodide/WebAssembly, establishing a pathway for high-fidelity multi-tier simulation. The resulting operational space is scientifically traceable and component-qualified for mission-scale pre-deployment experimentation.
\end{abstract}


\section{Introduction}
Underwater gliders enable persistent ocean observation by repeatedly diving and climbing while moving horizontally at low speed \cite{rudnick2016}. Their scientific return depends on the executed path in longitude, latitude, depth, and time, while currents, bathymetry, energy, and limited control authority complicate route execution \cite{leonard2010,subramani2017}. Field trials remain essential, but they are expensive, logistically constrained, and cannot expose several planning strategies to an identical, resettable ocean. Comparisons are also difficult to reproduce when methods use different environmental fields, vehicle assumptions, sensing rules, or scores.

We address this problem with a browser-native digital test range for pre-deployment experimentation. The browser hosts guided mission setup, data ingestion, planning, simulation, observation generation, evaluation, interactive 3D analysis, and artifact export. Static hosting reduces installation and administrative barriers on managed research and classroom machines. The work contributes: (1) an end-to-end plan-to-observation contract with planner/evaluator information separation; (2) a visible start-to-finish workflow for constructing a scientifically traceable 3D/4D operational experiment; (3) a common interface for manual, algorithmic, and imported learned planners; and (4) complementary qualification through controlled benchmarking, an authentic field-referenced audit, and native-to-browser execution of an established flight model. The primary contribution is a portable experimental contract through which heterogeneous planning methods can be compared from mission construction through executed observations and score.

\section{Related Work and Positioning}
Glider planning has been studied as current-aware, energy-aware, informative, and multi-objective optimization. Energy-optimal coastal planning accounts for time-varying currents \cite{subramani2017}; informative waypoint selection balances travel cost and observation value \cite{binney2013}, while broader observation strategies systematically analyze preferred glider trajectories \cite{smedstad2015expansion}. Multi-objective studies jointly consider forcing, sampling goals, and vehicle constraints \cite{lucas2019}, often extending into complex spatiotemporal multi-agent routing and sensor placement optimization \cite{holmberg2022stochastic,holmberg2024knowledge,holmberg2025knowledge}. OceanGNS demonstrates the operational value of a cloud portal that combines forecasts, bathymetry, glider data, planning, and pilot support \cite{oceangns2021}, operations which rely heavily on the efficient querying and mining of underlying spatio-temporal information systems \cite{ioup2007efficient,ladner2012mining}.

Vehicle-model and Observing System Simulation Experiment (OSSE) work addresses complementary layers. Published glider flight models represent buoyancy-driven motion and water-relative performance \cite{merckelbach2019}, while ocean OSSEs provide controlled references for comparing observing systems under explicit information policies \cite{halliwell2014}. Browser scientific computing provides a portable execution substrate: WebAssembly is a safe compilation target \cite{haas2017}, Pyodide exposes CPython and scientific Python in the browser, and client-side scientific applications can eliminate backend computation \cite{lun2023}. Virtual laboratories can also support active STEM experimentation when their models and learning objectives are explicit \cite{dejong2013virtual}. The gap addressed here is integration: a static client-side environment that makes domain construction, planner inputs, executed 4D motion, observations, scoring, and evidence boundaries inspectable under one contract.

\begin{figure*}[!t]
\centering
\definecolor{navybox}{RGB}{12, 28, 48}
\definecolor{boxborder}{RGB}{40, 60, 90}
\definecolor{dashline}{RGB}{120, 180, 200}

\resizebox{\textwidth}{!}{
\begin{tikzpicture}[
    font=\sffamily,
    mainbox/.style={
        rectangle, rounded corners, fill=navybox, text=white,
        text width=3.4cm, minimum height=2.7cm, inner sep=8pt,
        align=left, draw=boxborder, thick
    },
    whitebox/.style={
        rectangle, rounded corners, fill=white, text=black,
        text width=21.2cm, inner sep=10pt,
        align=left, draw=boxborder, thick
    },
    arrow/.style={->, >=Latex, very thick, navybox},
    link/.style={dashed, very thick, dashline}
]

    \node[mainbox] (n1) {
        \textbf{Episode inputs}\\[2mm]
        \fontsize{8.5}{10.5}\selectfont domain, clock, fleet, currents, environment roles
    };
    
    \node[mainbox, right=0.5cm of n1] (n2) {
        \textbf{Planner}\\[2mm]
        \fontsize{8.5}{10.5}\selectfont permitted fields only: bathymetry, masks, currents, prior, uncertainty, value
    };
    
    \node[mainbox, right=0.5cm of n2] (n3) {
        \textbf{Mission execution}\\[2mm]
        \fontsize{8.5}{10.5}\selectfont waypoints and dive policy produce current-advected 4D trajectory
    };
    
    \node[mainbox, right=0.5cm of n3] (n4) {
        \textbf{Observation operator}\\[2mm]
        \fontsize{8.5}{10.5}\selectfont records generated from executed trajectory
    };
    
    \node[mainbox, right=0.5cm of n4] (n5) {
        \textbf{Evaluator}\\[2mm]
        \fontsize{8.5}{10.5}\selectfont operational, dive-policy, and evaluator-only science metrics
    };

    \node[whitebox, below=0.6cm of n3] (n6) {
        \textbf{\large Mission View display and interaction branch}\\[1.5mm]
        \fontsize{9.5}{11.5}\selectfont Bathymetry, currents, science overlays, route geometry, trajectory inspection, camera, color, contours, and vertical exaggeration are visual interfaces. Display and camera state do not affect scientific results.
    };

    \draw[arrow] (n1) -- (n2);
    \draw[arrow] (n2) -- (n3);
    \draw[arrow] (n3) -- (n4);
    \draw[arrow] (n4) -- (n5);

    \draw[link] (n1.south) -- (n6.north -| n1.south);
    \draw[link] (n3.south) -- (n6.north -| n3.south);

\end{tikzpicture}
} 
\caption{Scientific computation chain and display boundary. Mission View visualizes frozen inputs and outputs, but camera, color, contours, and vertical exaggeration do not affect the planner, mission execution, observations, or evaluator.}
\label{fig:architecture}
\end{figure*}
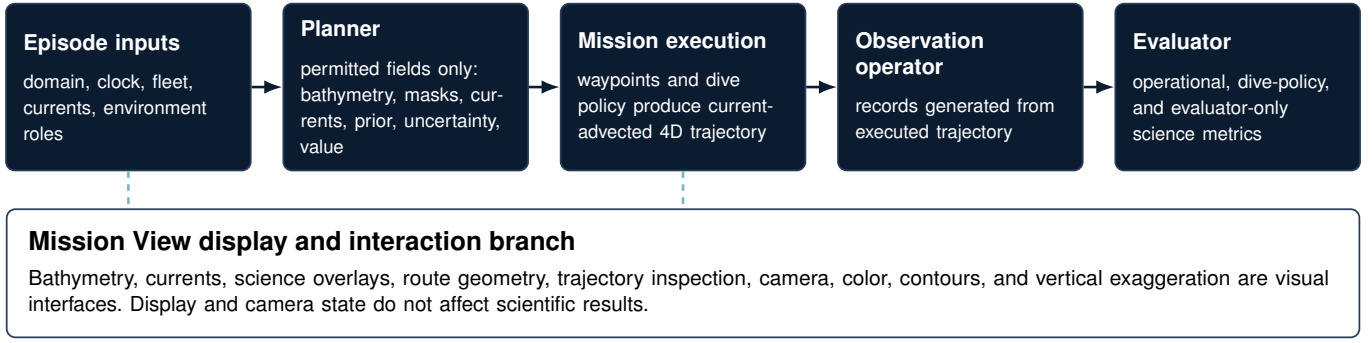

\section{Guided Digital Test-Range Workflow}
Figure~\ref{fig:architecture} summarizes the information boundary. The interface guides a user through \emph{Bathymetry $\rightarrow$ Currents $\rightarrow$ Science $\rightarrow$ Segments $\rightarrow$ Routes $\rightarrow$ Mission $\rightarrow$ Simulate $\rightarrow$ Validate}. Table~\ref{tab:walkthrough} identifies the principal choices and persisted object at each stage. A geographic bounding box defines the broad analysis domain. Mission-scoped bathymetry is extracted from the GEBCO\_2026 15 arc-second grid \cite{gebco2026}; valid-ocean, shallow/no-go, reachability, and optional task decomposition may further constrain operation. To remain static-host compatible, the client reads bounded raster windows to represent local terrain; the local WebGL view uses level-of-detail terrain, cached tiles, and packed height textures to support rapid 3D synthetic environment reconstruction \cite{abdelguerfi20013d}.

NetCDF, HDF5, and JSON adapters preserve environmental time/depth coordinates and source provenance, building on established frameworks for geographic data interchange, heterogeneous GIS interoperability, and web-service integration \cite{wilson2003geographical,tu2002achieving,tu2004integrating,chung2001geospatial}. Scientific depth arrays and masks are authoritative, whereas camera state and vertical exaggeration are display-only.

\begin{table*}[!t]
\centering
\caption{Guided mission-construction workflow, representative source/generator modes, and persisted scientific objects.}
\label{tab:walkthrough}
\scriptsize
\setlength{\tabcolsep}{2.7pt}
\begin{tabularx}{0.985\textwidth}{@{}>{\bfseries}p{0.098\textwidth}p{0.525\textwidth}X@{}}
\toprule
Stage / tab & Representative modes and user-controlled dimensions & Persisted scientific object \\
\midrule
Bathymetry & GEBCO mission-window extraction or frozen raster/GeoTIFF-derived domain; bounding box, grid resolution, reconstruction, and ocean/land/shallow/no-go definitions. & Mission depth grid and operational masks. \\
Currents & Off; controlled analytic generator; frozen current pack; NetCDF/HDF5; or historical HYCOM subset, with $u/v$, time/depth support, binding, and interpolation policy. & Provenanced current field available to planning and execution. \\
Science & Off; controlled front, eddy, plume, multiscale, or bathymetry-linked OSSE; or imported NetCDF/HDF5/environment pack, with reference, prior, uncertainty, objective, sampling value, and QC. & Planner-visible roles plus evaluator-only reference. \\
Segments & Off; sectors, survey corridors, target or reachable regions, or multi-vehicle responsibility partitions. & Explicit planner search-space/task contract. \\
Routes & Manual waypoints; direct/transect and A* baselines; or imported classical/learned-planner output, with acceptance and dwell semantics. & Ordered waypoint plan with provenance. \\
Mission & Clock, deployment/recovery, fleet, through-water speed, turn/vertical limits, dive policy, clearance, sensor, and observation schedule. & Reproducible mission specification. \\
Simulate & Timestep, common current-advection execution, completion/recovery, and constraint policy. & Executed 4D trajectory $(\lambda,\phi,z,t)$. \\
Validate/export & Common observation/QC and evaluator; operational, dive, and scientific metrics; artifact/checksum policy. & MissionResult plus plan, trajectory, observations, report, and dataset-ready records. \\
\bottomrule
\end{tabularx}
\vspace{-1.5mm}
\end{table*}

\begin{figure*}[!t]
\centering
\includegraphics[width=0.96\textwidth]{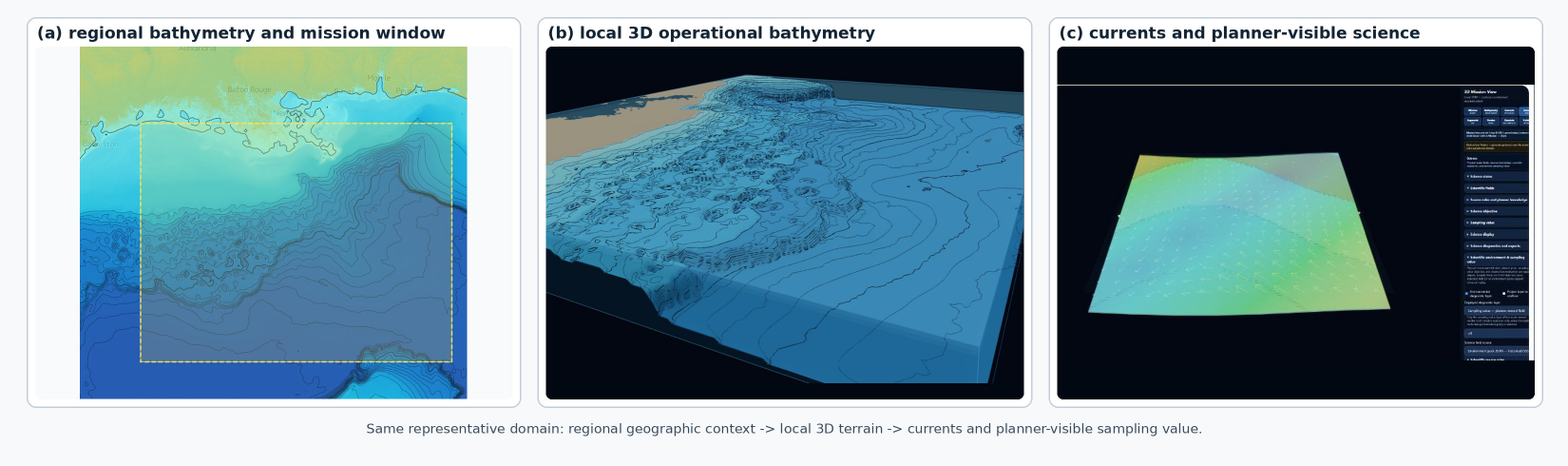}
\caption{From regional geographic context to a local 3D mission environment. Panel (a) shows Gulf Coast land, coastline, bathymetric relief, contours, and the selected mission window. Panel (b) shows the corresponding local 3D bathymetry, and panel (c) adds planner-visible sampling value and representative current vectors from one frozen mid-time/depth slice. The 3D relief is display-exaggerated for legibility; scientific depth arrays and calculations are unchanged.}
\label{fig:environment}
\end{figure*}

The environment contract separates a scalar reference, planner prior, uncertainty, declared objective, sampling value, and quality/provenance information. The hidden reference is evaluator-only. Currents and scalar fields may be controlled analytic episodes or frozen imported subsets with explicit time/depth support. Segmentation is optional: it defines regions, targets, corridors, or responsibility areas; routing defines intended guidance; simulation determines actual current-advected motion.

\subsection{Planning Modes and Dataset Generation}
Routes can be entered manually, generated by transparent built-in baselines, or imported as ordered waypoints from an external classical or learned planner. The same bathymetry, forcing, mission, vehicle, sensing, and evaluator remain fixed, preventing a new planner from silently changing the experimental problem. A browser-hosted model may execute through JavaScript or a Web Worker/Pyodide adapter, while an external method may emit the same plan artifact. Each run exports frozen inputs, plan, executed trajectory, observation records, metrics, report, and checksums. Repeated episodes and seeds can therefore be assembled into trajectory-imitation, planner-ranking, offline policy-evaluation, or other training/evaluation datasets. This paper establishes the reproducible data and evaluation contract needed to do so.

\section{Mission Execution and Evaluation}
The mission contract fixes deployment and recovery, clock, fleet, vehicle speed, dive policy, waypoint acceptance, dwell, clearance, observation schedule, and evaluation. The active benchmark engine advances the vehicle using mission-scale current-advection kinematics,
\begin{equation}
\mathbf{v}_{g}(t)=\mathbf{v}_{w}(t)+\mathbf{v}_{c}\!\left(x(t),y(t),z(t),t\right),
\label{eq:kinematics}
\end{equation}
where $\mathbf{v}_{g}$ is ground-relative velocity, $\mathbf{v}_{w}$ is commanded through-water velocity, and $\mathbf{v}_{c}$ is current sampled at the executed state. Heading and vertical-rate limits, a sawtooth dive policy, waypoint dwell, masks, and clearance rules are applied. The evaluator strictly scores the executed trajectory $(\lambda,\phi,z,t)$.

\begin{figure*}[!t]
\centering
\includegraphics[width=0.96\textwidth]{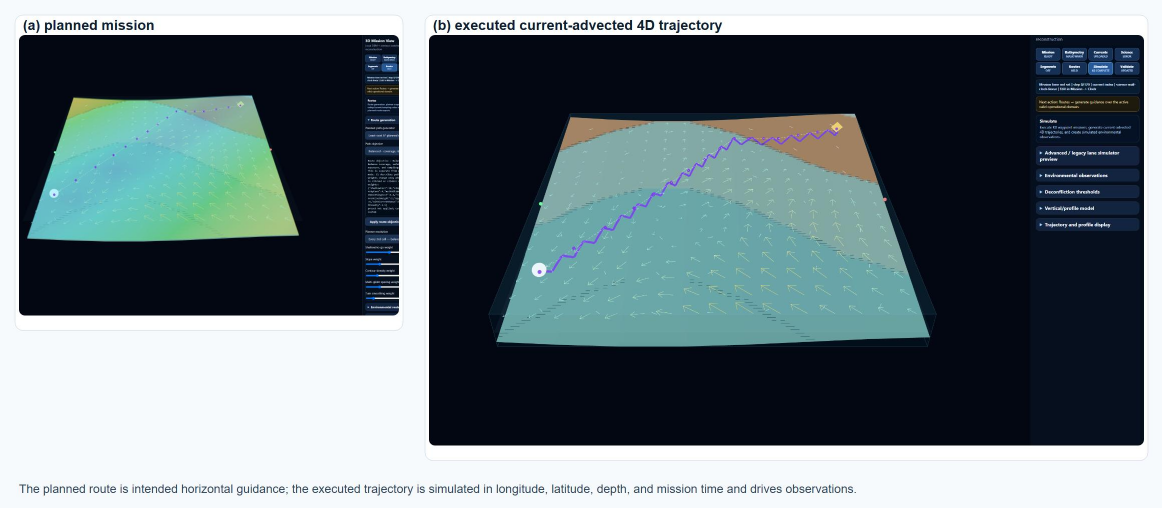}
\caption{Planned guidance and executed 4D motion for the representative \texttt{cross-shelf-front} science-aware A* run (fixed dive, seed 1). The current-advected trajectory, dive/climb cycle, observations, deployment, and recovery correspond to checksum \texttt{fnv1a32:2f0262b7}. Depth is exaggerated $20\times$ for display only.}
\label{fig:execution}
\end{figure*}

Observation candidates are scheduled from executed longitude, latitude, depth, and mission time. The reported OSSE uses noiseless pseudo-observations and a project-defined distance-weighted local-kernel update applied to the common prior; Barnes objective analysis supplies the methodological context for this spatial reconstruction \cite{barnes1964}. All methods use the same schedule, quality rules, and evaluator. Scientific performance is
\begin{equation}
S_{\mathrm{rec}}=1-\frac{\mathrm{RMSE}_{\mathrm{post}}}{\mathrm{RMSE}_{\mathrm{prior}}},
\label{eq:skill}
\end{equation}
with a prior-only control producing $S_{\mathrm{rec}}=0$. Operational, dive, and scientific metrics remain separate, and observation counts denote the total volume of accepted records.

To demonstrate the complete contract, we instantiate every stage in two frozen OSSE episodes, submit five planners to the same mission engine, and evaluate the resulting observations with one common reconstruction operator.

\section{Controlled Benchmark Design and Results}
Two deterministic episodes represent a cross-shelf front (CSF) and a depth-shear eddy (DSE). Each contains georeferenced bathymetry, masks, three current times and depths, a hidden scalar reference, prior, uncertainty, and sampling value. The one-glider mission uses through-water speed \SI{0.42}{m.s^{-1}}, a 30-minute step, and 121 states over 60 hours. Preflight found a \SI{74.772}{km} deployment-to-recovery separation and a 49.452-hour nominal no-current lower bound, making the original 12-hour design infeasible.

\begin{table}[t]
\centering
\caption{Transparent Track A planner definitions.}
\label{tab:planners}
\scriptsize
\setlength{\tabcolsep}{2.0pt}
\begin{tabularx}{\columnwidth}{@{}>{\bfseries}p{0.27\columnwidth}X@{}}
\toprule
Method & Principal behavior or cost \\
\midrule
Direct transit & Deployment and recovery waypoints only. \\
Cross-shelf transect & Three deterministic seed-dependent intermediate waypoints. \\
Bathymetry A* & Geodesic cost plus strong shallow-water and inverse-depth penalties. \\
Current A* & Geodesic cost plus opposing- and cross-current penalties from permitted mean currents. \\
Science A* & Geodesic cost minus $3500(V+0.15\sigma)$, where $V$ is planner-visible sampling value and $\sigma$ prior uncertainty; five intermediate waypoints and 1800 s midpoint dwell. \\
\bottomrule
\end{tabularx}
\vspace{-1mm}
\end{table}

\begin{table}[t]
\centering
\caption{Track A fixed-dive means over three deterministic seeds.}
\label{tab:main-methods}
\setlength{\tabcolsep}{2.2pt}
\scriptsize
\begin{tabular}{@{}llrr@{}}
\toprule
Episode & Method & Distance (km) & $S_{\mathrm{rec}}$ \\
\midrule
CSF & Direct transit & 89.497 & 0.966834 \\
CSF & Cross-shelf transect & 89.927 & 0.965819 \\
CSF & Bathymetry A* & 90.660 & \textbf{0.966927} \\
CSF & Current A* & 90.660 & \textbf{0.966927} \\
CSF & Science A* & 90.776 & 0.966904 \\
DSE & Direct transit & 88.857 & \textbf{0.966785} \\
DSE & Cross-shelf transect & 88.275 & 0.965481 \\
DSE & Bathymetry A* & 88.179 & 0.966698 \\
DSE & Current A* & 91.085 & 0.966675 \\
DSE & Science A* & \textbf{87.701} & 0.966076 \\
\bottomrule
\end{tabular}
\end{table}

All 54 ranking runs were numerically valid, operationally feasible, complete, and recovered, with zero hard violations. Completion and recovery occurred from 49 to 56 hours (mean 52.45 hours). 

\begin{figure*}[!t]
\centering
\includegraphics[width=0.88\textwidth]{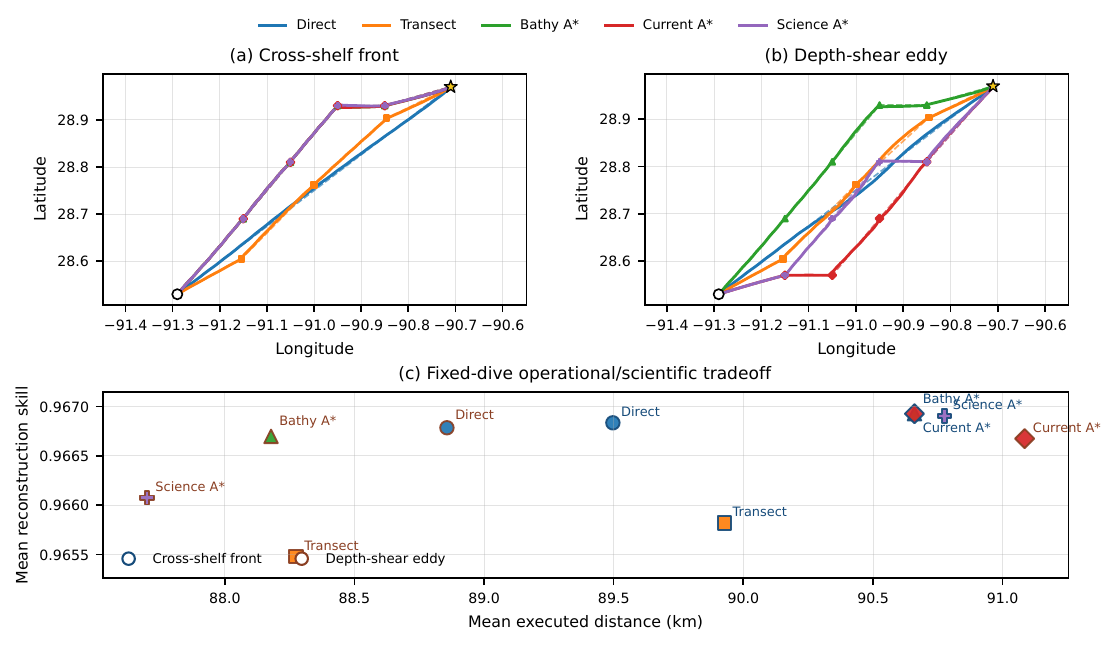}
\caption{Representative seed-1 Track A planned and executed trajectories for the two controlled OSSE episodes, with mean reconstruction skill versus executed distance. The clean geographic comparison complements the simulator-native 3D execution view in Fig.~\ref{fig:execution}; the shortest method is not consistently the highest-skill method.}
\label{fig:planner-comparison}
\end{figure*}

In CSF, direct transit is shortest, while bathymetry- and current-aware A* achieve the highest mean skill. In DSE, science-aware A* is shortest, while direct transit achieves the highest mean skill. These metric-dependent rank reversals are stable across the three seeds, but the absolute skill spread is small.

\begin{figure*}[!t]
\centering
\includegraphics[width=0.84\textwidth]{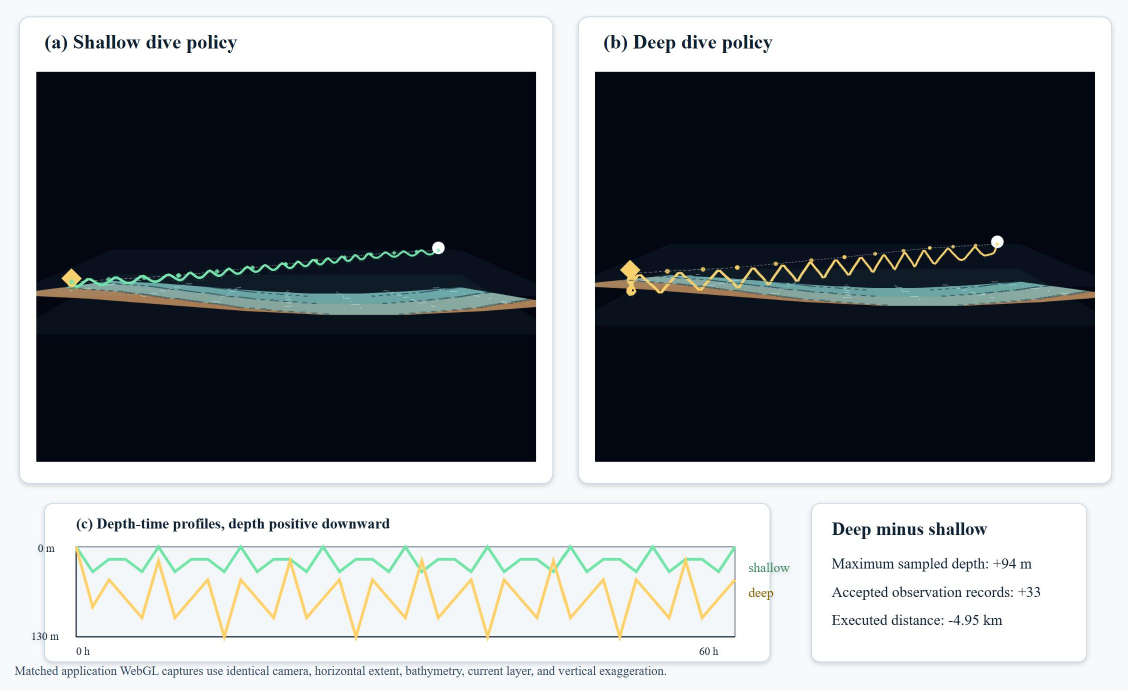}
\caption{Matched shallow and deep dive policies. Deep minus shallow changes are $+94$ m maximum sampled depth, $+33$ accepted observations for the displayed case, and $-4.95$ km executed distance.}
\label{fig:dive-policy}
\end{figure*}

Deep policies increase maximum sampled depth by 94 m, mean observation depth by about 48.7--49.3 m, and accepted observations by 31--34 records, while reducing executed distance by about 4.88--5.05 km. They also change current exposure and trajectory geometry. Skill changes are much smaller, approximately $-2.5\times10^{-4}$ to $2.9\times10^{-4}$. The evaluator is near saturated: 2112 lattice points, observation-to-lattice ratios of 0.091--0.114, and skill of 0.965--0.967. Smooth fields, a strong prior, noiseless observations, and broad kernel influence are the most consistent explanation. Results highlight descriptive operational-scientific tradeoffs among the planners.

\section{Scientific Qualification and Reproducibility}
The operational space is scientifically traceable and \emph{component-qualified} for simulation. Here, component-qualified means that individual data, computation, information-boundary, and browser-execution contracts have supporting evidence; this confirms bounded execution within the idealized simulator. Table~\ref{tab:qualification} separates verification, calibration, field-referenced auditing, reference-model portability, and benchmark inference.

\begin{table}[t]
\centering
\caption{Scientific qualification evidence and boundary.}
\label{tab:qualification}
\scriptsize
\setlength{\tabcolsep}{2.0pt}
\begin{tabularx}{\columnwidth}{@{}>{\bfseries}p{0.26\columnwidth}X@{}}
\toprule
Evidence & Outcome \\
\midrule
Known-answer verification & Current advection, heading/depth conventions, waypoint logic, information separation, and regression gates passed. \\
Mission calibration & 60 h horizon selected from preflight; 54/54 ranking missions completed and recovered. \\
Field-referenced audit & Real deployment exposed data and model defects; held-out endpoint reproduction was not accepted. \\
Browser model parity & Two native and two browser GliderFlight runs; eight fields deterministic with zero observed difference. \\
Benchmark inference & Common inputs and evaluator; three seeds support descriptive comparisons, not significance. \\
\bottomrule
\end{tabularx}
\end{table}

Known-answer cases cover zero and uniform current, spatially varying currents sampled at executed states, depth-shear reversal, heading/depth conventions, waypoint and clearance behavior, import/export stability, and browser/Node regression. In the passive-current oracle, an expected \SI{120}{m} displacement was reproduced within \SI{0.14}{m}, with current applied exactly once. These verify correct implementation and information boundaries.

To conduct the field-referenced audit, we selected a historical deployment from the public IOOS Glider Data Assembly Center (dataset identifier: \texttt{unit\_308-20220826T2012}). This dataset contained 512,157 records and 1,746 profiles reaching \SI{193.75}{m} \cite{unit308,ioosdac}. Public DAC $u/v$ were absent, so frozen HYCOM GOMu0.04 fields supplied forcing \cite{hycomgom}. Across eight held-out replay segments, HYCOM improved three endpoint predictions and worsened five; median paired error changed by $+\SI{1.16}{km}$. This outcome clarified the current engine's fidelity, indicating that exact field-track reproduction requires higher-resolution forcing and complex nonlinear vehicle modeling, which motivates a future transition toward multi-fidelity simulation.

To lay the groundwork for a high-fidelity simulation tier, a separate portability experiment evaluated the established GliderFlight 1.2.0 flight model. We executed unchanged GliderFlight 1.2.0 twice under native CPython and twice in a browser Web Worker through Pyodide 0.26.4 \cite{merckelbach2019,gliderflight120,pyodide0264}. Eight outputs were deterministic with zero observed native-browser difference under absolute tolerance $10^{-8}$ and relative tolerance $10^{-7}$. Native runs took 2.42--2.45 s; browser model execution took 4.20--4.29 s after initialization. While the current kinematic engine provides the rapid execution necessary for iterative planner benchmarking, this result verifies that the browser can reliably host the computationally heavier scientific software required for future high-fidelity validation.

Each certified run stores frozen inputs, plan, 4D trajectory, observations, report, and checksums; representative figures use seed 1 under a predeclared selection rule. Display/camera settings are excluded from scientific fingerprints. These artifacts support repeatable research comparisons and dataset production without requiring a native scientific stack.

\section{Accessibility, Use Modes, Limitations, and Conclusion}
\subsection{Manual, Algorithmic, and Learning-Based Workflows}
The common contract supports several ways to create a mission without changing how it is executed or scored. In a manual workflow, an operator or student authors waypoints, acceptance radii, and dwell behavior and compares the intended route with the current-advected result. In an algorithmic workflow, built-in or external planners receive the same permitted bathymetry, masks, currents, prior, uncertainty, sampling value, and mission constraints. In a learning-based workflow, a browser-hosted or external model emits the same ordered-waypoint artifact; the hidden reference remains unavailable and only the executed trajectory is observed and scored. This preserves a fair comparison between manual judgment, transparent baselines, optimization methods, and future ML policies.

The exported artifacts also define a dataset lifecycle. Frozen domain and environment packs describe the context; the route supplies the planner decision; the 4D trajectory and observation records describe consequences; operational, dive, and reconstruction metrics provide labels; and checksums preserve provenance. Batches across domains, current regimes, science objectives, dive policies, and seeds can support imitation of selected routes, prediction of mission outcomes, planner ranking, curriculum generation, or offline policy evaluation. The package can therefore produce dataset-ready records for training and testing, and the present study outputs synthetic, model-derived labels for this purpose.

\subsection{Interactive STEM Virtual Laboratory}
Static browser deployment lets researchers, students, and outreach participants inspect the same mission package on managed machines without installing Python, geospatial libraries, or a native simulator. The staged tabs and 3D Mission View support structured inquiry: learners can predict the effect of a current field, shallow mask, science objective, segmentation choice, route, or dive policy; change one factor while keeping the episode frozen; execute the mission; and compare the resulting trajectory, observations, and score. This design follows virtual-laboratory principles \cite{dejong2013virtual} and makes the distinction among a map, a plan, a physical trajectory, and a scientific result tangible. The present paper focuses strictly on demonstrating access, interaction, and reproducibility.

\subsection{Limitations and Future Validation}
The active engine is mission-scale kinematics rather than a field-validated nonlinear vehicle model; field replay did not reproduce authentic endpoints accurately; OSSE fields and observations are idealized; the experiment has three seeds; reconstruction skill is near saturated; no learned planner is evaluated; and the platform is not navigation-grade. Future work will couple the browser-hosted flight model to 4D forcing, validate held-out deployments across multiple missions, add noise and harder reference fields, evaluate manual/classical/learned planning modes under the same contract, and assess educational use.

\subsection{Conclusion}
The work establishes that an installation-free browser can host a complete and reproducible plan-to-observation benchmarking workflow and execute source-locked scientific software. The resulting digital test range constructs a traceable operating space, compares planning methods from intended route through executed observations and score, exports reusable artifacts, and provides a practical intermediate step between algorithm development and costly field deployment while broadening access to ocean-robotics experimentation.

\balance
\printbibliography
\end{document}